\documentclass[letterpaper, 10 pt, conference]{ieeeconf}

\IEEEoverridecommandlockouts                              

\usepackage[utf8]{inputenc}
\usepackage{amsmath}
\usepackage{graphicx}
\usepackage{multirow}
\usepackage{multicol}
\usepackage{scrextend}
\usepackage[table,xcdraw]{xcolor}
\usepackage{bm}
\usepackage{epsfig,psfrag,graphicx,color}
\usepackage{hyperref}
\usepackage[font=small,labelfont=bf]{caption}
\usepackage{comment}
\usepackage{colortbl}
\usepackage[percent]{overpic}

\usepackage{color}
\usepackage{xcolor,colortbl}

\usepackage{amssymb}
\usepackage{afterpage}
\usepackage{pdflscape}
\usepackage{enumerate}
\usepackage{booktabs}
\let\labelindent\relax
\usepackage{enumitem}

\usepackage{amsmath}
\usepackage{algorithm}
\usepackage[noend]{algpseudocode}

\usepackage{adjustbox}
\usepackage{array}

\usepackage[keeplastbox]{flushend} 

\usepackage{cite}

\usepackage[normalem]{ulem} 

\newcolumntype{R}[2]{%
	>{\adjustbox{angle=#1,lap=\width-(#2)}\bgroup}%
	l%
	<{\egroup}%
}

\newcolumntype{P}[1]{>{\raggedright\arraybackslash}p{#1}}

\graphicspath{{./figs/}}

\def\multiline#1{\shortstack[r]{#1}}

\newcommand{\doubtt}[1]{\textcolor{black}{#1}}

\newcommand{\ft}{{\small\textbf{[ft]}}}

\begin{document}

\title{Household Cloth Object Set: Fostering Benchmarking\\in Deformable Object Manipulation}

\author{Irene Garcia-Camacho$^1$, Júlia Borràs$^1$, Berk Calli$^2$, Adam Norton$^3$ and Guillem Alenyà$^1$\\
\thanks{This work receives funding from the Spanish State Research Agency through the projects BURG (CHIST-ERA - PCIN2019-103447) and CHLOE-GRAPH (PID2020-118649RB-l00). 
}
\thanks{${}^1$ Institut de Robòtica i Informàtica Industrial, CSIC-UPC, Barcelona, Spain
{\tt\small \{igarcia, jborras, galenya\}@iri.upc.edu}}%
\thanks{${}^2$  Worcester Polytechnic Institute, Massachusetts, USA
{\tt\small bcalli@wpi.edu}}%
\thanks{${}^3$  University of Massachusetts Lowell, USA
{\tt\small adam\_norton@uml.edu}}%
}

\maketitle

\begin{abstract}
Benchmarking of robotic manipulations is one of the open issues in robotic research. An important factor that has enabled progress in this area in the last decade is the existence of common object sets that have been shared among different research groups. However, the existing object sets are very limited when it comes to cloth-like objects that have unique particularities and challenges. This paper is a first step towards the design of a cloth object set to be distributed among research groups from the robotics cloth manipulation community. We present a set of household cloth objects and related tasks that serve to expose the challenges related to gathering such an object set and propose a roadmap to the design of common benchmarks in cloth manipulation tasks, with the intention to set the grounds for a future debate in the community that will be necessary to foster benchmarking for the manipulation of cloth-like objects. Some RGB-D and object scans are collected as examples for the objects in relevant configurations and shared in \url{http://www.iri.upc.edu/groups/perception/ClothObjectSet/HouseholdClothSet.html}
\end{abstract}



\section{Introduction}
\label{sec:introduction}

Benchmarking in the context of manipulation has gained a lot of attention in the last decade because it is crucial to enable progress. However, benchmarking for manipulation remains a challenge in the community due to several reasons. Large number of robot embodiments make repeatability and comparison between results difficult. The variety of strategies to solve a task and the high level of intertwinement between perception, planning and control make it difficult to describe the building blocks that need to be solved to resolve a task. Also, the need to demonstrate results with real robot executions make it very costly to gather data that can be easily reused or to generate databases where the results can be tested in a similar way as computer vision has done.

\begin{figure}
    \centering
    \includegraphics[width=\linewidth]{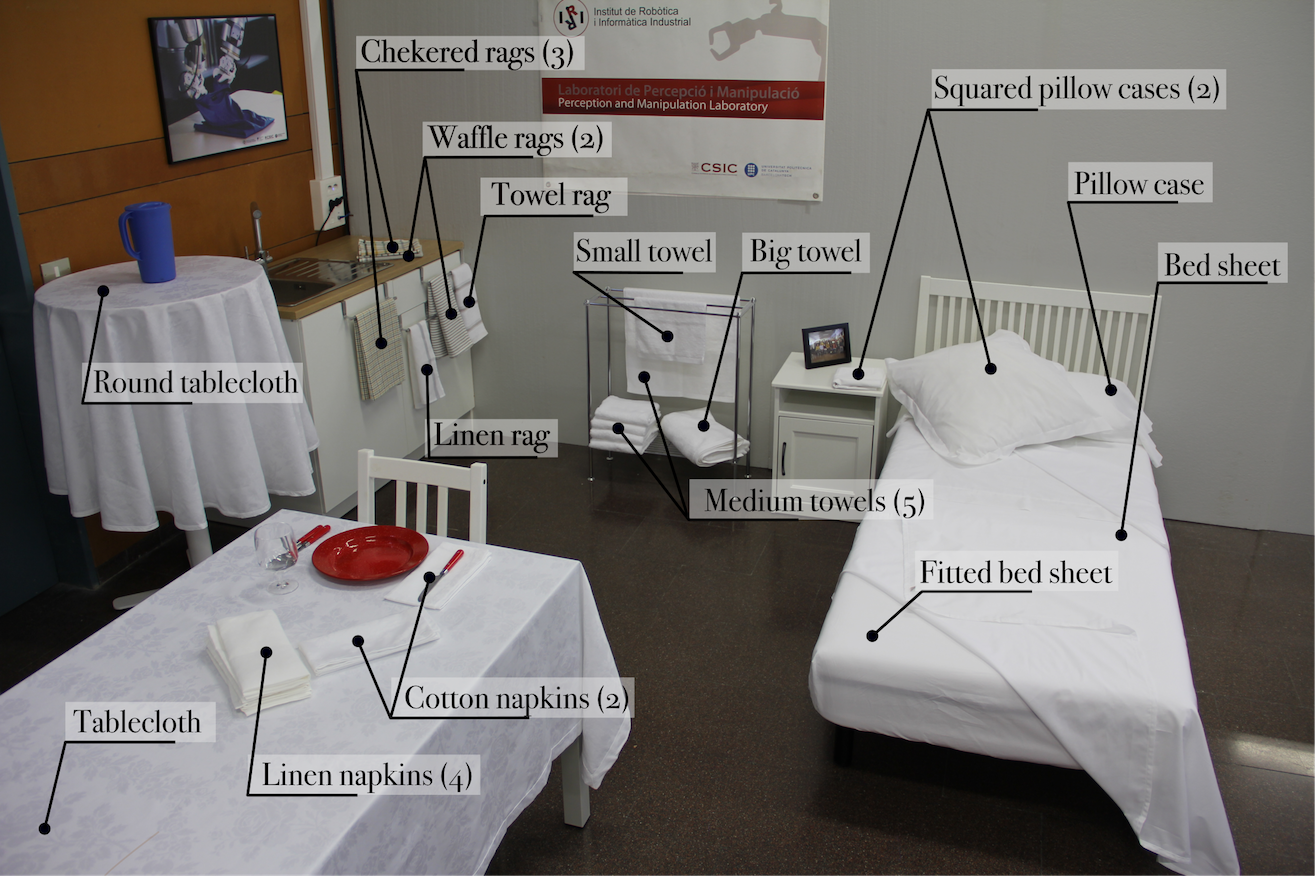}
    \caption{The household cloth object set. Includes objects of very different sizes, repetitions, different shapes and different types of cloth}
    \label{fig:cloth_set}
\end{figure}

A big step towards standarization was done in \cite{calli2015} with the introduction of the YCB object set with physical objects together with other data like object models and templates to define protocols and benchmarks for different tasks. The considerably wide spread of use of this  object set has allowed progress towards standarization and comparision of results. In this work, we present a first step towards the extension of the YCB object set towards highly deformable cloth-like objects (Fig. \ref{fig:cloth_set}). Our  objective is to sparkle a discussion in the community of robotic cloth manipulation towards the definition and creation of a cloth-like object set with several categories: household items, dressing items, cloth-type samples and rigid cloth-related items. In this work, we concentrate on the first category, household items, together with a proposal of relevant information like weight and sizes, and object models in relevant configurations. Household items are the simplest category in terms of geometry but are challenging enough to identify the problems and issues that need to be solved. 

One of the main challenges when creating standarized object sets is the continuity of the stock of a particular item. That was already an issue for the YCB object set, but the span of the life cycle for a clothing item is even shorter, making it very difficult to consider the option of providing a shopping list like the one proposed in \cite{leitner2017}. Previous works on benchmarking cloth manipulation also indicated a shopping list for the objects used \cite{garcia-camacho2020benchmarkbimanual}, but links are not longer available. Alternatively, for T-shirts, \cite{garcia-camacho2020benchmarkbimanual} just indicated the measures. However, different T-shirts of the same size can behave very differently depending on the fabric properties. 
For this reason, we have acquired a large number of object sets to be distributed among the research community. To develop the rest of the object categories, we plan to extend the collaboration among more researchers in the field to maximise the acceptance and use of the set.

\begin{table*}[!htp]
\vspace{6pt}
\centering
\caption{Literature object and cloth data sets.}
\scriptsize
\begin{tabular}{l|r|r|r|r|r|r|c}
\multirow{3}{*}{}	&\multirow{3}{*}{\textbf{Dataset}}	&\multirow{3}{*}{\textbf{Year}}	&\multirow{3}{*}{\textbf{Objects}}	&\multirow{3}{*}{\textbf{Data type}}	&\multirow{3}{*}{\textbf{Purpose}}	& \textbf{Physical}	&	\\
&	&	&	&	&	&\textbf{objects}	& \textbf{Textile}	\\
&	&	&	&	&	&\textbf{available?}	& \textbf{objects?}	\\\hline \hline

\multirow{3}{*}{1} & \multirow{3}{*}{YCB\cite{calli2015ycb}} & \multirow{3}{*}{2015} & \multirow{3}{*}{77} & RGB-D images &  & \multirow{3}{*}{\textbf{Yes}} & \multirow{3}{*}{2} \\
& & & & High-res images & Grasping& & \\
& & & & 3D mesh models & & & \\\hline

\multirow{3}{*}{2} & \multirow{3}{*}{ACRV \cite{leitner2017}} & \multirow{3}{*}{2017} & \multirow{3}{*}{42} & RGB-D images &  & \multirow{3}{*}{\textbf{Yes}} &  \\
& & & & High-res images & Grasping & & 2 \\
& & & & 3D mesh models & & & \\\hline

3	& Household objects\cite{choi2009objects}	&2009	&43	& -	&Grasping	&\textbf{Yes}	& 4	\\\hline


\multirow{3}{*}{4}	&\multirow{4}{*}{KIT database \cite{kasper2012KIT}}	&\multirow{3}{*}{2012}	&\multirow{3}{*}{100}	&RGB images	&Recognition,	&\multirow{3}{*}{No}	&\multirow{3}{*}{No}	\\
&	&	&	&3D mesh models	&localization and	&	&	\\
&	&	&	&Grasp annotation?	&manipulation	&	&	\\\hline

\multirow{2}{*}{5}	&\multirow{2}{*}{VisGraB \cite{kootsra2012VisGrasB}}	&\multirow{2}{*}{2012}	&\multirow{2}{*}{18}	&Stereo images	&\multirow{2}{*}{Grasping}	&\multirow{2}{*}{No}	&\multirow{2}{*}{No} \\
&	&	&	&3D models	&	&	&	\\\hline

\multirow{2}{*}{6} & \multirow{2}{*}{BigBird \cite{Singh2014BigBird}} & \multirow{2}{*}{2014} & \multirow{2}{*}{127} & 3D mesh models & \multirow{2}{*}{Recognition} & \multirow{2}{*}{No} & \multirow{2}{*}{No} \\
 & & & & high-res images & & & \\\hline 

7 &\multirow{2}{*}{CLUBS\cite{novkovic2019clubs}}	&\multirow{2}{*}{2019}	&\multirow{2}{*}{85}	&\multirow{2}{*}{RGB-D dataset}	&Segmentation,	&\multirow{2}{*}{No}	&\multirow{2}{*}{8}	\\
&	&	&	&	&classification and detection	&	&	\\\hline

8 &DexNet \cite{mahler2019Dexnet}	& 2017	&40 &3D mesh model	&Grasp evaluation	&No	& 1	\\\hline

\multirow{3}{*}{9}	&\multirow{3}{*}{GraspNet \cite{fang2019GraspNet}}	&\multirow{3}{*}{2020}	&\multirow{3}{*}{88}	&RGB-D images &\multirow{3}{*}{Grasp evaluation}	&\multirow{3}{*}{No}	&\multirow{3}{*}{No}	\\
	&	&	&	&Grasp pose annotation	&	&	&	\\
	&	&	&	&3D mesh models	&	&	&	\\\hline
	
10 & Warehouse P\&P \cite{rennie2016}	&2016	&25	&RGB-D dataset	&3D pose estimation	&No	&No	\\\hline


11 & Rothlin \cite{rothling2007}	&2007	&21	&3D models	&Grasping	&No	& No	\\\hline

12 & DefGraspSim \cite{huang2021defgraspsim} & 2021 & 34 & Mesh models &  Grasp evaluation& No & No \\\hline

\rowcolor[gray]{0.9}13 & Yamazaki \cite{yamazaki2013}	&2013	&21	&RGB images	&Classification & No & \textbf{Yes}	
\\\hline
 
\rowcolor[gray]{0.9}14 & Mariolis \cite{mariolis2013}	& 2013 &12	& \multiline{RGB images synthetic dataset}& Shape-matching & No & \textbf{Yes}\\ 
	\hline

\rowcolor[gray]{0.9}15 &Doumanoglou	\cite{doumanoglou2014} & 2014	& 6	& Depth	&Recognition & No & \textbf{Yes} \\\hline	

\rowcolor[gray]{0.9}16 & Glasgow's Database	\cite{aragoncamarasa2013glasgows} &2013	&16	&stereo-pair RGB images	& 3D point cloud	&No & \textbf{Yes} 
\\\hline

\rowcolor[gray]{0.9}17 &Willimon	\cite{willimon2013} &2013	&7	& RGB-D images	& \multiline{Feature detection and Classification}& No & \textbf{Yes} \\	
\hline

\rowcolor[gray]{0.9}18 & Ramisa	\cite{Ramisa_eaai14} &2014	&6	&RGB-D images	&Feature extraction & No & \textbf{Yes}	
\\\hline

\rowcolor[gray]{0.9}19 & Corona	\cite{Corona_PR18} &2018	&4	&Depth	& Classification and Grasp location & No & \textbf{Yes}\\


\bottomrule
\end{tabular}
\label{tab:literature_datasets}
\end{table*}

Cloth objects are complex to describe. The textile industry has many tools to define the properties of cloth to a very fine degree of precision, probably not needed  for robotics. However, the cloth tags only provide information of the yarn material, that is not enough to deduce the dynamic properties of a piece of cloth. 
The work in \cite{longhini2021textile}  proposed a taxonomy to classify cloth type based on material and on the fabrication method, and they developed a classifier that could distinguish between woven vs. knitted samples of polyester, wool and cotton materials. 
In this work, we annotate the dimension of each cloth item in rest position but also under pulling forces along the edges and also diagonally (shear) to capture the different deformability capacities of each item. In addition, in the website we provide microscopic images of the fabrics, whose inspection was used in \cite{longhini2021textile} to get the ground truth of the fabrication method of each analysed fabric.

\section{Related Work}\label{sec:soa}

Progress in computer vision has been possible mainly thanks to the creation of datasets used for  perception tasks such as detection, classification and pose estimation, providing images, point clouds, and meta-data of different collections of rigid objects. This line of work has been extended to the study of simulated grasping of rigid objects with other types of data such as 3D mesh models and grasp annotations. Table \ref{tab:literature_datasets} summarises some of the object sets found in the literature that have been used for robotic grasping and manipulation. 
The vast majority of these datasets are not intended to be used in real experimental work and do not provide the necessary information for its physical acquisition. However, having the physical objects available has been shown very relevant to allow progress and fair comparison of results. 
For instance, the YCB object set \cite{calli2015ycb} is now widely established and many groups test their algorithms using it. It consists of a wide range of objects to allow different manipulations. It includes two textile objects: a tablecloth and a T-shirt. In addition, as it was designed for benchmarking manipulation, they ensured the physical acquisition of objects through the distribution of several sets among the research community. Alternatives include the works \cite{choi2009objects, leitner2017} that provided shopping lists to purchase objects and included some textiles as shirts, pants, gloves or socks. Nevertheless, these three works together only sum 7 different textile objects. Also, these works provided a database including RGB-D images and 3D mesh models for all the objects except for those that do not have a determinate shape, i.e., textile objects.

\begin{table*}
\vspace{6pt}
    \centering
    \footnotesize
    \caption{List of objects}
    \setlength\tabcolsep{3pt} 
    \begin{tabular}{l|l|l|l|l|l|l|l}
         \textbf{\rotatebox[origin=c]{90}{Category}} & \textbf{Name} &  \textbf{\rotatebox[origin=c]{90}{Quantity}} &
         \textbf{\rotatebox[origin=c]{70}{Approx. Dimens. (m)}} & \textbf{\rotatebox[origin=c]{70}{Approx. Weight (Kg)}} & \textbf{\rotatebox[origin=c]{70}{Length elasticity}} & \textbf{\rotatebox[origin=c]{70}{Width elasticity}} & \textbf{\rotatebox[origin=c]{70}{Diagonal elasticity}} \\ \hline \hline

         \multirow{3}{*}{\textbf{\rotatebox[origin=c]{90}{Bath.}}}  
         & Small Towel &1 & $0.3\times 0.5$ & 0.080 & 5\% & 4\% & 17\% \\ \cline{2-8}
         & Med. towel & 5 & $0.5\times 0.9$ &  0.22 & 4\% & 2\% & 13\%\\ \cline{2-8}
         & Big towel & 1 & $0.9\times 1.5$ & 0.64 & 5\% & 4\% & 12\%\\ \hline
        
         \multirow{4}{*}{\textbf{\rotatebox[origin=c]{90}{Bedroom}}}
         & Bedsheet & 1 & $1.6\times 2.8$ & 0.5 & 2\% & 2\% & 9\%\\ \cline{2-8}
         & Fitted bedsheet & 1 & \small{$0.9\times 2\times 0.3$}  & 0.38 & 3\% & 3\% & 5\%\\ \cline{2-8}
         & Sq. pillowcase & 2 & $0.7\times 0.7$ & 0.12 & 2\% & 3\% & 9\%\\ \cline{2-8}
         & Rect. pillowcase & 1 & $0.45\times 1.1$ & 0.13 & 2\% & 1\% & 5\%\\ \hline
         
         \multirow{4}{*}{\textbf{\rotatebox[origin=c]{90}{Dinning}}} 
         & Rect. tablecloth & 1 & $1.7\times 2.5$ & 0.75  & 2\% & 1\% & 8\% \\\cline{2-8}
         & Round tablecloth & 1 & $\varnothing2$ & 0.56 & 1\% & -   & - \\ \cline{2-8}
         & Cotton napkin & 2 & $0.5\times 0.5$ & 0.05 & 2\% & 1\% & 13\% \\ \cline{2-8}
         & Linen napkin & 4 & $0.5\times 0.5$ & 0.05 & 2\% & 2\% & 16\% \\  \hline 
         
         \multirow{4}{*}{\textbf{\rotatebox[origin=c]{90}{Kitchen}}} 
         & Towel rag & 1 & $0.5\times 0.7$ & 0.05 & 4\% & 4\% & 17\% \\ \cline{2-8}
         & Linen rag & 1 & $0.5\times 0.7$ & 0.05 & 1\% & 2\% & 10\% \\ \cline{2-8}
         & Waffle rag & 2 & $0.5\times 0.7$ & 0.10  & 4\% & 4\% & 23\% \\ \cline{2-8}
         & Chekered rag & 3 & $0.5\times 0.7$ & 0.08 & 4\% & 7\% & 17\% \\ \hline
         
    \end{tabular}\\
    \label{tab:object_set}
\end{table*}

Similar to rigid object datasets, there are some cloth-related sets in the literature, some of them collected in the grey final rows of \autoref{tab:literature_datasets}. There are more sets from the literature of computer vision, but we have focused only on the literature for robotic applications.
The majority of these works create a small object set of garments, trying to span different types and shapes, for testing their perception and manipulation approaches. However, this object sets are not intended to be used for others and therefore no physical information of the objects is provided. Some examples of this are \cite{Doumanoglou_ActivePerceptRandForest_icra14, yamazaki2013, mariolis2013} which purpose was to demonstrate the effectiveness of their approaches in classifying and recognising different clothes.

To the best of our knowledge, no object set rich in a variety of textile objects exists to the moment, not to say that there are no cloth-like sets that can be acquired physically for testing manipulations with real robotic systems.


\section{The cloth object set}\label{sec:cloth_set}

In this section, we describe the cloth object set with the reasoning of the selection and describing the possibilities that offer. In contrary to the works seen in the literature (with the exception of \cite{calli2015ycb}), this cloth set is intended to be distributed among different research institutions, offering an easy access for its use in real manipulation applications for benchmarking purposes.

\subsection{Object Set} 

In this set we concentrate on household objects, leaving dressing items and other categories for future work, with the aim of offering a wide variety of objects for this category. Our aim is to stablish the criteria for selecting the items and the kind of information that is needed. An overview of the proposed object set can be seen in \autoref{fig:cloth_set}, in  \autoref{tab:object_set} and in more detail in the related website\footnote{\url{http://www.iri.upc.edu/groups/perception/ClothObjectSet/HouseholdClothSet.html}}.

The object set selection consists of objects that can be found in any house and are used to perform a wide range of household chores tasks such as folding, pilling, table setting or bed making. As this object set is designed for benchmarking manipulation it covers sizes from 30cm up to 200cm to provide different complexity of manipulations and allowing to be used with different sized workspaces. It also includes different textile fabrics and textures, which offers variability in the textile properties such as rigidity, elasticity and roughness, what affects aspects as the dynamics of the object under contacts. Also, the variety on size and fabric types offer objects with different weights, suitable for several end-effector payloads. Table \ref{tab:object_set} picks up the physical properties of the objects as their dimension, weight and \% of elasticity under tension. Since due to the manufacturing process the dimensions and weight of the same objects may vary a little, we have measured the real dimensions of the acquired objects and computed the difference to the labelled value, finding that for the majority of the objects there is an error between 0.5\% and 1\% but for some rags this error is of up to a 3\%. 

In the textile market, the supply is highly variable over time, making it difficult for a product to remain stable during the seasons and even less from one year to the next. This has been the main selection criterion. For this reason, we have prioritized choosing basic products in terms of fabric and colour, as there are the objects that last longer available.

The object set provides: 
\begin{itemize}
 \item Very \textbf{different sized objects} of the same type, that allows to compare strategies for manipulating small and larger objects that require different robotic workspace sizes.
 \item It has \textbf{repeated objects} like the 5 medium towels, to allow tasks like piling/unpiling.
 \item It has \textbf{repeated not equal objects} like rags or napkins with the same size but different fabric properties, allowing the possibility to compare one method on different textiles which can provide insight on adaptability on fabric dynamics and tips to improve the performance.
 \item It has very \textbf{large sized objects of different materials and weights}, like the bedsheet, the big towel or the tablecloths. These type of big objects require  to develop novel grasping strategies that do not use perception due to its large size, e.g. once a corner is grasped, extra manipulations are necessary to locate and reach the second corner, like tracing the edge or with re-grasps.
 \item It includes a \textbf{round shaped object} with which it cannot make use of the corner location strategy for manipulation and requires strategies based on geometries for example for folding tasks.
 \item It has items with \textbf{double layer} like pillowcases similar as in T-shirts. These objects may require different manipulation strategies for properly handle them.
 \item It also has objects with \textbf{complex features} like the fitted bedsheet, requiring to implement force strategies for the manipulation of the adjustable hem, or the tubular pillowcase which involve a complex tasks like fit a pillow in the pillowcase without slipping from the other side. 
\end{itemize}

\subsection{Initial setup and maintenance}

\begin{figure}[tb]
\vspace{6pt}
    \centering
    \includegraphics[width=0.8\linewidth]{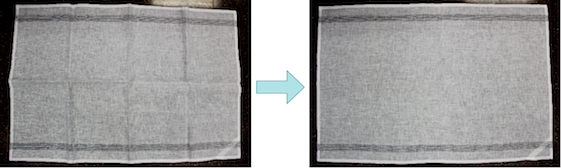}
    \caption{Example of wrinkle removal through ironing}
    \label{fig:wrinkled_rag}
\end{figure}

\begin{figure}[tb]
    \centering
    \includegraphics[width=0.7\linewidth]{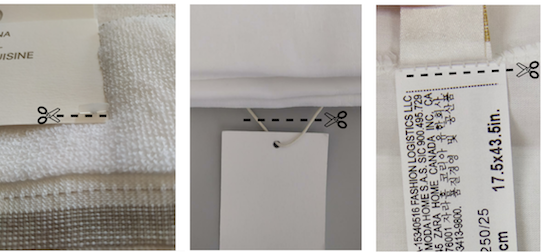}
    \caption{Examples of cloth tags that need to be removed before use}
    \label{fig:cut_tags}
\end{figure}

This cloth set aims to be used in several research institutions working on deformable object manipulation for benchmarking purposes. For this reason, once the cloth set is acquired it is necessary to follow some indications in order to maintain the objects as far as possible in the same conditions, as textiles properties can be greatly modified depending on the treatment and use you give to them. Thus, we defined a list of do's and dont's:

\begin{itemize}
    \item Washing: Washing modifies the properties of clothes in different manners depending on the type of washing machine, water properties, washing products, etc. what makes it difficult to define a common protocol for washing. For this reason, it is not permitted to wash the objects.
    \item Ironing: Ironing is allowed in order to remove lines and wrinkles (see Fig. \ref{fig:wrinkled_rag}) to avoid errors as they influence the cloth state and in their behaviour in some task executions, such as placing or folding, tending to appear bendings in the parts where there are folding lines. . 
    \item Dyeing: If desired, it is possible to dye some relevant parts of the cloth as edges or corners to ease perception. To do so, use some textile markers that do not damage or modify the fabric properties of the cloth. For example, tape or stickers fixed on the corners will increase the rigidity of the cloth, but textile liquid colorants or marker pens will not.
    \item Cut tags: All objects come in a packaging with three different type of tags (see Fig. \ref{fig:cut_tags}). The two first ones must be removed by cutting the plastic strip and the cord in order to properly handle the objects. For some manipulations (e.g. edge tracing) or perception applications it is also recommended to cut the third tag cutting the label flush with the hem, taking care not to cut the seam that joins it or damage the fabric (see Fig. \ref{fig:cut_tags}c). The hanger is useful for setting the cloths on a common hanging configuration so we recommend not cutting it but if necessary it is allowed to do so.
\end{itemize}

\section{Benchmarking guidelines}
  
In this section, we explore the challenges faced when designing a benchmark in cloth manipulation and provide guidelines that can serve as a roadmap to do so, using the present cloth set. 

\vspace{-0.1cm}
The structure of current robotic manipulation benchmarks consists of a setup description, a task description and a definition of the assessment measures \cite{calli2015}. 
In the case of benchmarking cloth manipulation, the setup description includes the hardware (e.g. robotic system, perception sensors, etc) and the textile objects that have to be used as well as the initial configurations from where to start. The task description explains the main actions of the task to be evaluated and the expected outcome. Finally, the evaluation metrics can consider only the final result of the task or could also provide assessment on the substates of the task (e.g. success in grasping the first corner in a spreading task) ~\cite{borras2020graph}. As it has been said, the present cloth set is intended to be used as the objects of future cloth manipulation benchmarks. Regarding the hardware, we consider that its restriction to specific ones is counterproductve since textiles already give the sufficient challenges, so any robotic system with grasping capabilities can be employed. The following sections of this paper attend the rest of the aspects: In section \ref{sec:initial_conf} relevant configurations of textiles are identified providing detailed protocols and in Section \ref{sec:tasks} cloth manipulation tasks are proposed as tasks for benchmarking with a description of the main actions and expected outcomes, as well as a discussion of possible assessment measures.

\subsection{Protocols for initial configurations}\label{sec:initial_conf}

As clothes can adopt infinite states, contrary to rigid objects, it is necessary to identify some configurations in order to properly handle them. Some of these configurations are relevant because they are usually initial or final states in cloth manipulation tasks. We describe here these configurations and 
clear protocols to set the objects in these states.

The relevant configurations identified in the literature are:
\begin{enumerate}
    \item[\ft] \textbf{Flat}: Completely spread in a flat surface.
    \item[\small\textbf{[Fd]}] \textbf{Folded}: Folded in halfs three times. 
    \item[\small\textbf{[Hco]}] \textbf{Hanging corner}: Hanging grasped by a corner. 
    \item[\small\textbf{[H2co]}] \textbf{Hanging two corners}: Hanging grasped by two adjacent corners, having the object flat in the air.
    \item[\small\textbf{[Hh]}] \textbf{Hanging hanger}: Placed in a wall hanger. All the towels and rags have a hanging strip.
    \item[\small\textbf{[Cr]}] \textbf{Crumpled}: Any other configuration not represented.  
    \item[\small\textbf{[Pl]}] \textbf{Pile}: Folded cloths stacked one on top of the other.
    \item[\small\textbf{[Fit]}] \textbf{Fitted}: Pillowcases with padding, fitted bedsheet in bed.

\end{enumerate}

We propose protocols for setting the cloths in these initial configurations. This is a necessary point for common replicable research for benchmarking purposes, as the starting configuration of the cloth greatly influences on the required manipulations to accomplish a determinate task.

\subsubsection{Flat state}
This state is the most basic configuration of a cloth, useful for its recognition, and necessary as a starting state for performing common tasks such as folding. A perfect placed cloth is the one that has been flattened removing wrinkles and bends. We suggest having the cloth flat with the hanger faced on the top side as the standard flat state \textbf{[Ft]}. 

\subsubsection{Folded state}
Although folded could be considered a simple state for cloths, there are many possible ways to fold even a simple rectangular cloth. For instance, a cloth can be folded in halfs or in three parts, joining the adjacent corners of the shorter or the longer edge, etc. 

\begin{figure}[tb]
\vspace{6pt}
    \centering
    \includegraphics[width=\linewidth]{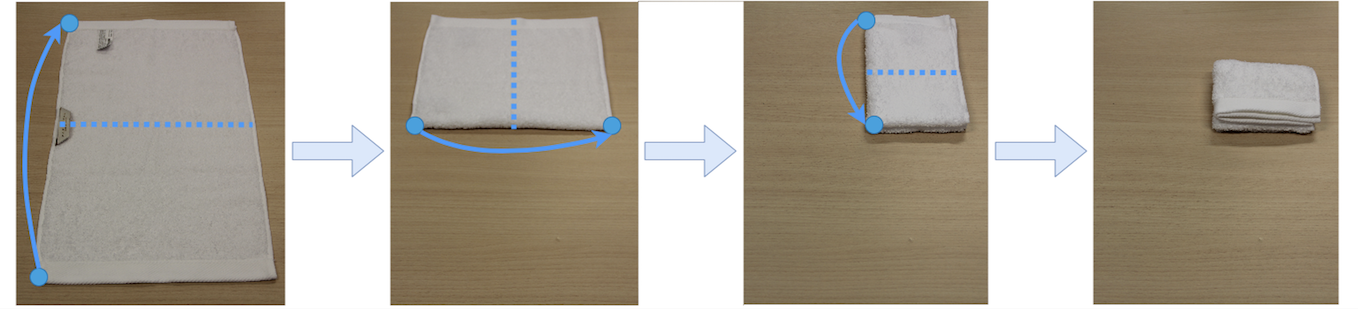}
    \caption{Three fold protocol sequence.}
    \label{fig:folding_protocol}
\end{figure}

\begin{figure}[bt]
    \centering
    \includegraphics[width=\linewidth]{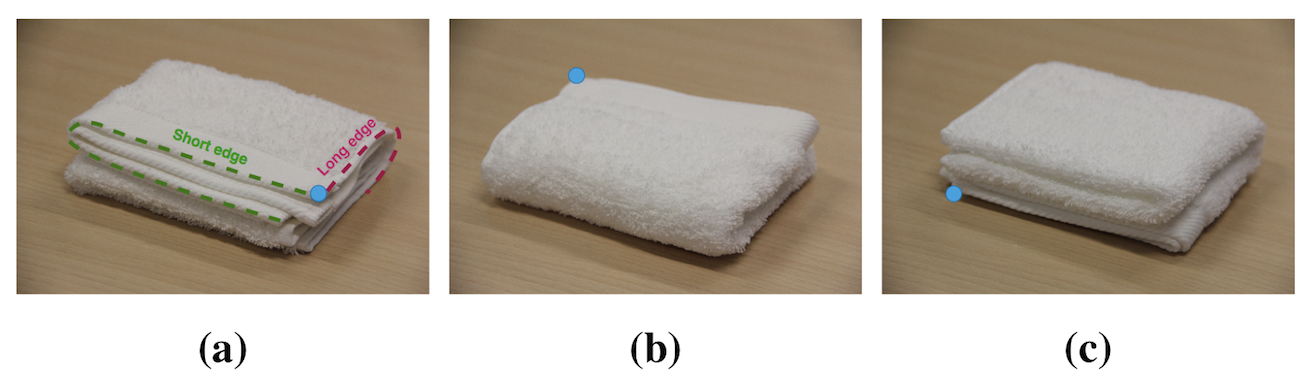}
    \caption{Three examples of increasing complexity according to the rotation of the folded cloth.}
    \label{fig:degrees_complexity}\vspace{-0.2cm}
\end{figure}

The folding \textbf{[Fd]} protocol described in this section consists on a three-fold pipeline (see Fig. \ref{fig:folding_protocol}). This procedure results in a cloth placed on the table with the corner visible on top of the cloth and facing the shorter edge (see Fig. \ref{fig:degrees_complexity}a).

\begin{enumerate}[label=(\alph*)]
    \item Place flat the cloth on a table, with the labels on the top side.
    \item Fold in half by the long edge and join the corners.
    \item Repeat steps 1 and 2 two times.
\end{enumerate}

Observe that the folded cloth can be rotated in order to increase the complexity of grasping, triggering the need to regrasp the cloth to access the corner (Fig. \ref{fig:degrees_complexity}).

This protocol can be performed for most of the objects, but due to their smaller and larger sizes, for the small towel it is sufficient to perform two folds and for the rectangular tablecloth and bedsheet it is recommended five folds.

The round tablecloth and fitted bedsheet require specific folding protocols due to their shape or characteristics (step-by-step videos are in the related webpage).

\subsubsection{Hanging one corner state}
In any cloth manipulation task it is a necessary step to grasp first one of the corners and lift the cloth. Some works in literature use this state to locate other relevant parts to grasp as edges or other corners \cite{Corona_PR18}. The hanging \textbf{[Hco]} protocol defined consists on placing the cloth flat as in the flat protocol, grasp one of the corners and lift it up with a vertical motion. 

\subsubsection{Hanging two corner state}
In the same way, a cloth hanging by two corners is a usual state when manipulating clothes, as it is a relevant configuration for recognition purposes and is considered the final state of an unfolding in the air. The hanging \textbf{[H2co]} protocol consists on grasping two adjacent corners of the shorter edge of the cloth and lift it up until having the cloth completely flat in the air. 

\subsubsection{Hanging by hanger state}
In addition to the previous two states, another easy way to normalise the hanging state of a cloth is \textbf{[Hh]} by using the hanger strip attached to one of the edges, found in all the towels, napkins and rags. 

\subsubsection{Crumpled state}
This configuration is very difficult to define as can correspond to an infinite number of possible states of the cloth. This makes extremely difficult to define clear protocols that ensure repeatibility. One attempt to define a protocol for crumpled \textbf{[Cr]} can be found in \cite{garcia-camacho2020benchmarkbimanual}.

\subsubsection{Pile}
It is common to find stacks of folded clothes in human environments as it is a simple and clean way to organise and store several items. This state is a derivative of the folded state, so the protocol to set several cloth in a piled state \textbf{[Pl]} consists on following the folding protocol with five different objects (e.g. five medium towels or five rags) and place them on a table one on top of the other having each of them the same orientation.

\subsubsection{Fitted}
The fitted state \textbf{[Fit]} has been defined for two specific objects of the cloth set according to their real applications: fitted bedsheet and pillowcases. The first one consists on having the fitted bedsheet correctly placed on a mattress, i.e. place the bedsheet flat on a single bed and fit the four elastic hems of the corners under the corners of the bed until the bedsheet is fully stretched. On the other side, the pillowcases are meant to cover pillows of 60x60cm and \doubtt{100x45cm} for the squared and the rectangular, respectively. In order to properly fit the squared pillowcase, this should have the entire padding inside with the flap closure covering it. On the other hand, the rectangular pillowcase has tubular shape with both sides open, so when adjusting the pillow filling inside it should be in the center, without coming out through any of the openings.

\subsection{Cloth manipulation tasks}\label{sec:tasks}

We propose some cloth manipulation tasks that can be implemented with the present object set, that provide many manipulation opportunities. The list of tasks include unimanual and bimanual manipulation tasks, which spans different complexity levels in manipulation, including some commonly studied tasks in manipulation research as folding or pick\&place but also some uncommon ones as making beds or inserting pillows in covers, giving the opportunity to study the development of ubiquitous tasks in our everyday life.

\vspace{-0.2cm}
\begin{multicols}{2}
\begin{itemize}
    \item Folding
    \item Unfolding
    \item Piling/Unpiling
    \item Pick \& Place
    \item Spreading
    \item Bed making
    \item Pillowcase fitting
\end{itemize}
\end{multicols}
\vspace{-0.2cm}

\begin{table}
\vspace{6pt}
    \centering
    \footnotesize
    \caption{Objects vs Tasks}
    \setlength\tabcolsep{3pt} 
    \begin{tabular}{l|c|c|c|c|c|c}\toprule 
         \multirow{2}{*}{\textbf{Object}} & \multirow{2}{*}{\textbf{Task}} & \multicolumn{5}{c}{\textbf{Initial states}} \\
         & & Ft & Fd & Hco & H2co & Cr\\ \hline \hline
         
         & \textbf{Folding} & X & & X & X & X\\ \cline{2-7}
         Small Towel & \textbf{Unfolding} & X & X & X & & X \\ \cline{2-7}
         Big Towel & Pick\&Place & & X & & & X \\ \cline{2-7} 
         & Spreading &  & X & X & X & X \\ \hline
        
          & Folding & X & & X & X & X \\ \cline{2-7}
          & Unfolding & X & X & X & & X \\ \cline{2-7}
         Medium Towel & Pick\&Place & & X & & & X \\ \cline{2-7} 
         (x5) & Spreading & & X & X & X & X \\ \cline{2-7} 
         & \textbf{Piling/Unpiling} & & X & & & \\ \hline
         
         & Folding & X & & X & X & X \\ \cline{2-7}
         Sq. pillowcase & Unfolding & X & X & X & & X \\ \cline{2-7}
         Rect. pillowcase & Pick\&Place & & X & & & X \\\cline{2-7}
         & \textbf{Fitting} & X & & X & X & \\\cline{2-7}
         & Spreading & & X & X & X & X \\  \hline
         
         & Folding & X & & X & X &  \\ \cline{2-7}
         Fitted bedsheet & Pick\&Place & & X & & & X \\ \cline{2-7}
         & \textbf{Fitting} & X & & X & X & \\\cline{2-7}
         & Spreading & & & X & X & X \\ \hline 
         
         \multirow{1}{*}{Rect. tablecloth} & Folding & X & & X & X & X \\ \cline{2-7}
         Round tablecloth & Unfolding & X & X & X & & X \\ \cline{2-7}
         Beedsheet & \textbf{Spreading} & & X & X & X & X \\ \cline{2-7}
         & Pick\&Place & & X & & & X \\ \hline
         
         Cotton napkin & Folding & X & & X & X & X \\ \cline{2-7}
         Linen napkin & Unfolding & X & X & X & & X \\ \cline{2-7}
         Towel rag & Spreading &  & X & X & X & X \\ \cline{2-7}
         Waffle rag & Unfolding & X & X & X & & X \\ \cline{2-7}
         Linen rag & \textbf{Pick\&Place} & X & & & & X \\ \cline{2-7}
         Chekered rag & Piling/Unpililng & & X & & & \\
       
    \bottomrule     
    \end{tabular}
    \label{tab:objectsvstasks}\vspace{-0.3cm}
\end{table}

\begin{table*}[tb]
\vspace{6pt}
\centering
\caption{Objects vs Dataset}\label{tab:dataset}
\scriptsize
\begin{tabular}{l|r|r|r|r|r|r}\toprule
\multirow{2}{*}{\textbf{Object}}	&\multicolumn{6}{c}{\textbf{Cloth configuration}}	\\\cline{2-7}
&Ft	&Fd	&Hco	&H2co	&Pile	&Fitted	\\ \hline \hline

Small towel&-	&3D model	&	-&	-&-	&-	\\
Big towel	&-	&F-RGBD	&F-RGBD	&F-RGBD	&-	&-	\\ 
&Z-RGBD	&Z-RGBD	&-	&-	&-	&-	\\ \hline

Medium towel&-	&3D model	&-	&-	&3D model	&-	\\
x5	&-	&F-RGBD	&F-RGBD	&F-RGBD	&F-RGBD	&-	\\
&Z-RGBD	&Z-RGBD	&-	&-	&-	&-	\\ \hline

Squared pillowcase&-	&3D model	&-	&-	&-	&3D model	\\
Rectangular pillowcase &-	&F-RGBD	&F-RGBD	&F-RGBD	&-	&F-RGBD	\\
&Z-RGBD	&Z-RGBD	&-	&-	&-	&-	\\ \hline

Bedsheet	&-	&3D model	&-	&-	&-	&-	\\
Fitted bedsheet	&-	&F-RGBD	&-	&-	&-	&F-RGBD	\\
Rect. tablecloth	&-	&Z-RGBD	&-	&-	&-	&-	\\
Round tablecloth	&	&	&	&	&	&	\\ \hline

	&-	&3D model	&-	&-	&3D model	&-	\\
Napkins and rags	&-	&F-RGBD	&F-RGBD	&F-RGBD	&F-RGBD	&-	\\
	&Z-RGBD	&Z-RGBD	&-	&-	&-	&-	\\

\bottomrule

\end{tabular}
\end{table*}

In Table \ref{tab:objectsvstasks} we list the objects that can be used in each of the tasks, relating them with the initial configurations from which they can be started. For the sake of clearness, we have grouped some objects as some of them can be used in the same combination of tasks and initial states. Highlighted in bold there are tasks that are most natural for each group of objects (e.g. Fitting for the pillowcases).

\subsubsection{Folding}\label{sec:folding}

Due to the relevance of this task, there is already a benchmark related to this task in  \cite{garcia-camacho2020benchmarkbimanual}.

\textbf{Task definition:} Grasp two adjacent corners and manipulate the cloth to join the contrary adjacent, folding the cloth by half. This process should be repeated for the desired number of folds. Its final state will outcome in \textbf{[Fd]}.

\textbf{Setup description:} This task can be performed with any of the objects presented. With respect to the initial configurations, \textbf{[Fd]} and \textbf{[[Ft]} states are the most used when performing this task but \textbf{[Hco]} and \textbf{[H2co]} can be considered as initial configurations to simplify the number of manipulations required to perform the task by reducing its pipeline execution.

\textbf{Evaluation and metrics:} To assess the performance of this task, it will be necessary to measure the initial and final area of the cloth and compare them. The metrics will be defined according to the folding approach, if the folding protocol of Section \ref{sec:initial_conf} is followed, the final area should be $1/8$ of the entire area of the cloth. Also, measuring the area of each of the three folds is useful in order to track possible errors that could be hidden between the consecutive folds. The area after each of the folds should be $1/2$ of the area from which it started. In addition, a wrinkle or bend estimation will be useful to provide a quality of the folds.

\subsubsection{Unfolding}

\textbf{Task definition:} Unfolding is a useful task for setting the objects to its canonical form for recognition purposes or as a preparation step for continuing with other tasks as folding or spreading. This task basically consists on grasping a corner of the cloth and then locating and grasping the second corner until having the cloth spread in the air. Its final state corresponds to \textbf{[H2co]}. 

\textbf{Setup description:} This task can be executed with any of the objects, but we suggest the exception of the fitted bedsheet since due to their elastic hems its quite difficult to mantain the cloth completely flat so it will be complex to assess whether it is successfully unfolded or not. The initial configurations could be folded \textbf{[Fd]}, crumpled \textbf{[Cr]}, flat \textbf{[Ft]} or hanging by one corner \textbf{[Hco]}.

\textbf{Evaluation and metrics:} A method to assess the success of this task could be to use a technique of shape detection using a RGB camera in order to match it to the predefined geometry of the object. Also, the distance of the end effectors to the corners could be measured, having more score the closer it is, since this could ease the next action (e.g. folding or placing).

\subsubsection{Pick\&Place}

\textbf{Task definition:} This is the most common and simpler task performed with rigid objects, but in the case of textiles it is not as trivial as it seems. Picking and placing crumpled clothes is usually performed for laundry applications, where the configuration of the cloth is not important, but it does not offer much manipulation challenges. In contrary, picking and placing folded items suppose to grasp and move the cloth maintaining the same configuration without unfolding it, what usually requires bimanual manipulations or the implementation of more complex strategies as dynamic motions in the placement. 

\textbf{Setup description:} According to the task description, the initial states for this task could be both crumpled \textbf{[Cr]} or folded \textbf{[Fd]}. Any of the objects of the cloth set could be used for its implementation, having more or less complexity according to the weight and fabric dynamics of the object.

\textbf{Evaluation and metrics:} The assessment of this task consists on the placement error, that includes the position and orientation error comparing it to the expected goal. In addition, in the case of pick\&place of folded items the evaluation should also consider the configuration of the placed cloth (i.e. if it has been unfolded or not).

\subsubsection{Pilling/Unpilling}

\textbf{Task definition:} This task is a derivative of the previous one, where the objects are whether initially piled or the placement goal is to place them one on top of the other.

\textbf{Setup description:} The objects more suitable for this task are the ones repeated, that is to say, the five medium towels or the rags. The initial configuration can be with the individual objects folded \textbf{[Fd]} or with all of them piled \textbf{[Pl]}.

\textbf{Evaluation and metrics:} This task is one of the most complex to assess. Scoring can include the pose of each of the piled objects, the quality of the pile stability, the quantity of wrinkles and bends of each piled item or how much each of them has been unfolded, etc. Some automatic tool could consist on a point cloud or model matching of the final pile with its corresponding 3D model provided.

\subsubsection{Spreading}
 
\textbf{Task definition:} This task consists on placing the objects into the flat state \textbf{[Ft]} on top of a table. Depending on the object, it will either be completely inside the table or will cover entirely the table as in the case of the tablecloths.
all the way up or stick out the sides
This will either be all the way up or stick out the sides

\textbf{Setup description:} Any of the objects can be used. 
All initial configuration can suite with this task except the flat state \textbf{[Ft]}, which corresponds to the task outcome.

\textbf{Evaluation and metrics:} The rotation and translation errors of the spread cloth is a good assessment metric for this task. In addition, for small/medium objects, some automatic tool to detect bends or wrinkles can also be used performing shape matching, obtaining a continuous scoring as in \cite{irene2021placing}. For larger items we can find an example in 
\cite{garcia-camacho2020benchmarkbimanual} where it is proposed a detailed benchmark for the tablecloth sprading task from the task description to the definition of specific evaluation metrics.

\subsubsection{Bed making and Pillowcase fitting}

 Some objects of the set allow to implement cloth manipulation tasks which have not been much studied in literature due to their high complexity. These two tasks require bimanual robotic systems working in a wide workspace due to the size of the clothes as the bedsheets. They also require complex manipulations as for example force-position controls for fitting the elastic hem of the fitted bedsheets, synchronized dynamic motions for spreading and inserting a pillow in a pillowcase without it goes out from the other side.

\section{Data set}\label{sec:dataset}

In addition to the distribution of the physical set, we created a very simple database of the objects in order to provide an overview of the individual objects in different formats and configurations. This includes an RGB-D dataset of all the objects in different configurations as well as 3D mesh models of some folded and piled items. In addition, microscopic images are also available for a more detailed view of the yarn type of each object. All these data can be downloaded from the paper website. The purpose of this database is not for being used as training as it does not have enough data, but as testing data for some computer vision and simulation approaches as feature detection, classification, object scene initialization, etc. Table \ref{tab:dataset} presents a summary of the type of data provided for each object and configuration.

\begin{figure}[tb!]
    \centering
    \includegraphics[width=0.8\linewidth]{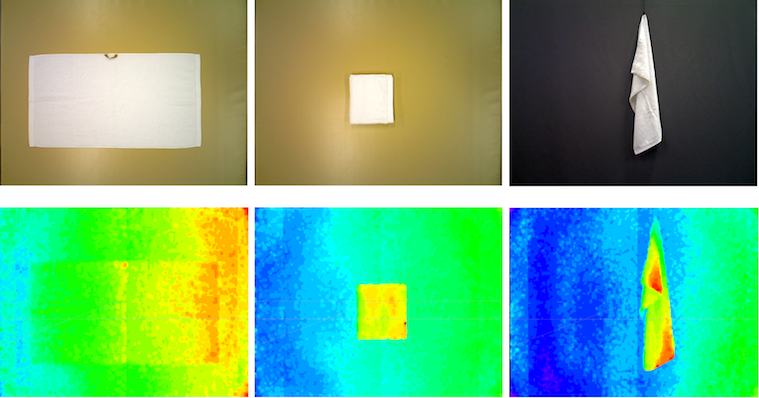}
    \caption{Color and depth data (top and bottom rows, respectively) for the medium towel for states \textbf{[Ft]}, \textbf{[Fd]} and \textbf{[Hco]} (from left to right).}
    \label{fig:rgbd_images}\vspace{-0.5cm}
\end{figure}

To capture color and depth images we have used an Asus Xtion Pro Live camera placed on tripods to take frontal and zenithal images (check the website for details). For the zenithal images, clothes where initially placed flat on a table and then folded in three folds following protocols in \autoref{sec:initial_conf}. Color and depth images are taken for the initial flat configuration and for each of the folds performed. For the frontal images, clothes are placed hanging from a corner to give an overview of their deformability and size and compare between the objects.

\begin{figure}[tb]
\vspace{6pt}
    \centering
   \includegraphics[width=0.7\linewidth]{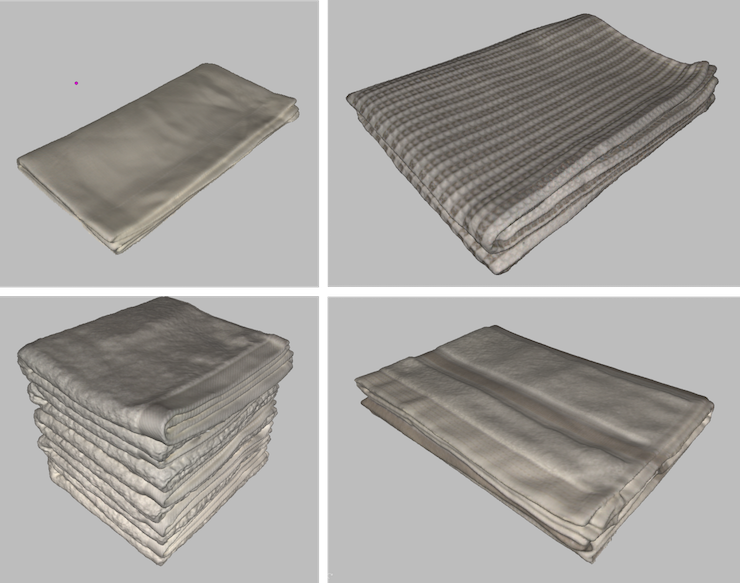}
    \caption{3D model scan examples, from top-left to bottom-right, the linen napkin, the waffle rag, a pile of towels and the towel rag.}
    \label{fig:scans}\vspace{-0.2cm}
\end{figure}

Figure \ref{fig:scans} shows some examples of the 3D models that are also available for all folded objects as well as the piled medium towels and stacked rags. To obtain the models we used the Artec Eva scanner, and as with the RGB-D images, all the clothes are folded following the protocol in Section \ref{sec:initial_conf}, and stacked in same orientation for the pilled scans. These models can be used for scene initialization in a simulator.

In addition to the RGB-D dataset and 3D mesh models of the objects, we took microscopic views (check related website for download) of all the textiles for identifying the type of yarn as proposed in \cite{longhini2021textile}.
This images have been taken with a Sensofar plu2300 and have an augmentation of x5, but in contrary to \cite{longhini2021textile}, microscopic images are not conclusive for determining the type of knit of the clothes.

\section{Conclusions}

We believe that the community can benefit from a well-stablished physical object set that serves to create benchmarks.
With this aim, a carefully selection of textile household objects found in our daily life is presented, spanning a variety of cloth types, sizes and characteristics that allow multiple robotic tasks possibilities of different natures. We also offered solutions to some of the most relevant challenges presented when benchmarking cloth manipulation such as protocols to set the clothes in relevant configurations. The paper also presents a simple dataset provided to show the potential of the cloth set besides its use for physical manipulation.

How to design benchmarks for robotic manipulation, and particularly for textile manipulation, is an intense discussion tackled in several active projects and among different institutions, because it is very complex.
We belief that the distribution of the object set will set the grounds for a collaboration between these institutions to enlarge the present object set with more textile objects to perform other applications such as dressing (e.g. T-shirts, trousers, etc) or with cloth-type samples to cover as many textile materials as possible.

\bibliographystyle{ieeetr}
\bibliography{clothManipulation} 

\begin{thebibliography}{10}

\bibitem{calli2015}
B.~Calli, A.~Walsman, A.~Singh, S.~Srinivasa, P.~Abbeel, and A.~M. Dollar,
  ``Benchmarking in manipulation research: Using the yale-{CMU}-berkeley object
  and model set,'' {\em IEEE Robot. Autom. Mag.}, vol.~22, no.~3, pp.~36--52,
  2015.

\bibitem{leitner2017}
J.~Leitner, A.~W. Tow, N.~S{\"u}nderhauf, J.~E. Dean, J.~W. Durham, M.~Cooper,
  M.~Eich, C.~Lehnert, R.~Mangels, C.~McCool, {\em et~al.}, ``The acrv picking
  benchmark: A robotic shelf picking benchmark to foster reproducible
  research,'' in {\em IEEE Int. Conf. on Rob. and Autom.}, pp.~4705--4712,
  IEEE, 2017.

\bibitem{garcia-camacho2020benchmarkbimanual}
I.~{Garcia-Camacho}, M.~{Lippi}, M.~C. {Welle}, H.~{Yin}, R.~{Antonova},
  A.~{Varava}, J.~{Borras}, C.~{Torras}, A.~{Marino}, G.~{Alenyà}, and
  D.~{Kragic}, ``Benchmarking bimanual cloth manipulation,'' {\em IEEE Robot.
  and Autom. Let.}, vol.~5, no.~2, pp.~1111--1118, 2020.

\bibitem{calli2015ycb}
B.~Calli, A.~Singh, A.~Walsman, S.~Srinivasa, P.~Abbeel, and A.~M. Dollar,
  ``The ycb object and model set: Towards common benchmarks for manipulation
  research,'' in {\em Advanced Robotics (ICAR), 2015 Int. Conf. on},
  pp.~510--517, 2015.

\bibitem{choi2009objects}
Y.~S. Choi, T.~Deyle, T.~Chen, J.~D. Glass, and C.~C. Kemp, ``A list of
  household objects for robotic retrieval prioritized by people with als,'' in
  {\em 2009 IEEE Int. Conf. on Rehab. Rob.}, pp.~510--517, 2009.

\bibitem{kasper2012KIT}
A.~Kasper, Z.~Xue, and R.~Dillmann, ``The kit object models database: An object
  model database for object recognition, localization and manipulation,'' {\em
  The Int. Journal of Rob. Res.}, vol.~31, 07 2012.

\bibitem{kootsra2012VisGrasB}
G.~Kootstra, M.~Popović, J.~Jorgensen, D.~Kragic, H.~Petersen, and N.~Krüger,
  ``Visgrab: A benchmark for vision-based grasping,'' {\em Paladyn}, vol.~3,
  2012.

\bibitem{Singh2014BigBird}
A.~Singh, J.~Sha, K.~S. Narayan, T.~Achim, and P.~Abbeel, ``Bigbird: A
  large-scale 3d database of object instances,'' {\em 2014 IEEE Int. Conf. on
  Rob. and Autom. (ICRA)}, pp.~509--516, 2014.

\bibitem{novkovic2019clubs}
T.~Novkovic, F.~Furrer, M.~Panjek, M.~Grinvald, R.~Siegwart, and J.~Nieto,
  ``Clubs: An rgb-d dataset with cluttered box scenes containing household
  objects,'' {\em The International Journal of Robotics Research (IJRR)},
  vol.~38, no.~14, pp.~1538--1548, 2019.

\bibitem{mahler2019Dexnet}
J.~Mahler, J.~Liang, S.~Niyaz, M.~Laskey, R.~Doan, X.~Liu, J.~Aparicio, and
  K.~Goldberg, ``Dex-net 2.0: Deep learning to plan robust grasps with
  synthetic point clouds and analytic grasp metrics,'' in {\em Rob. Science and
  Syst.}, 07 2017.

\bibitem{fang2019GraspNet}
H.-S. Fang, C.~Wang, M.~Gou, and C.~Lu, ``Graspnet-1billion: A large-scale
  benchmark for general object grasping,'' in {\em 2020 IEEE/CVF Conference on
  Computer Vision and Pattern Recognition (CVPR)}, pp.~11441--11450, 2020.

\bibitem{rennie2016}
C.~Rennie, R.~Shome, K.~E. Bekris, and A.~F. De~Souza, ``A dataset for improved
  rgbd-based object detection and pose estimation for warehouse
  pick-and-place,'' {\em IEEE Robot. and Autom. Let.}, vol.~1, no.~2,
  pp.~1179--1185, 2016.

\bibitem{rothling2007}
F.~Rothling, R.~Haschke, J.~J. Steil, and H.~Ritter, ``Platform portable
  anthropomorphic grasping with the bielefeld 20-dof shadow and 9-dof tum
  hand,'' in {\em 2007 IEEE/RSJ Int. Conf. on Intel. Rob. and Syst.},
  pp.~2951--2956, 2007.

\bibitem{huang2021defgraspsim}
I.~Huang, Y.~Narang, C.~Eppner, B.~Sundaralingam, M.~Macklin, T.~Hermans, and
  D.~Fox, ``Defgraspsim: Simulation-based grasping of 3d deformable objects,''
  2021.

\bibitem{yamazaki2013}
K.~Yamazaki and M.~Inaba, ``Clothing classification using image features
  derived from clothing fabrics, wrinkles and cloth overlaps,'' in {\em 2013
  IEEE/RSJ Inte. Conf. on Intel. Rob. Syst.}, pp.~2710--2717, 2013.

\bibitem{mariolis2013}
I.~Mariolis and S.~Malassiotis, ``Matching folded garments to unfolded
  templates using robust shape analysis techniques,'' in {\em Int. Conf. on
  Comp. Analysis of Images and Patterns}, 08 2013.

\bibitem{doumanoglou2014}
A.~Doumanoglou, A.~Kargakos, T.-K. Kim, and S.~Malassiotis, ``Auton. active
  recognition and unfolding of clothes using random decision forests and prob.
  planning,'' in {\em 2014 IEEE Int. Conf. on Robot. and Autom. (ICRA)},
  pp.~987--993, 2014.

\bibitem{aragoncamarasa2013glasgows}
G.~Aragon-Camarasa, S.~B. Oehler, Y.~Liu, S.~Li, P.~Cockshott, and J.~P.
  Siebert, ``Glasgow's stereo image database of garments,'' 2013.

\bibitem{willimon2013}
B.~Willimon, I.~Walker, and S.~Birchfield, ``Classification of clothing using
  midlevel layers,'' {\em ISRN Robotics}, vol.~2013, 03 2013.

\bibitem{Ramisa_eaai14}
A.~Ramisa, G.~Aleny{\`a}, F.~Moreno-Noguer, and C.~Torras, ``Learning rgb-d
  descriptors of garment parts for informed robot grasping,'' {\em Engin.
  Applic. of Artif. Intel.}, vol.~35, pp.~246--258, 2014.

\bibitem{Corona_PR18}
E.~Corona, G.~Aleny\`{a}, T.~Gabas, and C.~Torras, ``Active garment recognition
  and target grasping point detection using deep learning,'' {\em Pattern
  Recognition}, vol.~74, pp.~629--641, 2018.

\bibitem{longhini2021textile}
A.~Longhini, M.~C. Welle, I.~Mitsioni, and D.~Kragic, ``Textile taxonomy and
  classification using pulling and twisting,'' {\em arXiv e-prints},
  pp.~arXiv--2103, 2021.

\bibitem{Doumanoglou_ActivePerceptRandForest_icra14}
A.~{Doumanoglou}, A.~{Kargakos}, T.~{Kim}, and S.~{Malassiotis}, ``Auton.
  active recognition and unfolding of clothes using random decision forests and
  prob. planning,'' in {\em IEEE Int. Conf. Robot. Autom.}, 2014.

\bibitem{borras2020graph}
J.~{Borràs}, G.~{Alenyà}, and C.~{Torras}, ``Encoding cloth manipulations
  using a graph of states and transitions,'' {\em arXiv:2009.14681}, 2020.

\bibitem{irene2021placing}
I.~Garcia-Camacho, J.~Borras, and G.~Alenya, ``Benchmarking cloth manipulation
  using action graphs: an example in placing flat,'' {\em IROS 2021 Workshop on
  Benchmarking of robotic grasping and manipulation}, 2021.

\end{thebibliography}

\end{document}